\documentclass[sigconf,screen]{acmart}

\renewcommand\footnotetextcopyrightpermission[1]{} 
\AtBeginDocument{%
  }

\usepackage{bbding}
\usepackage{makecell}
\usepackage{amssymb}
\usepackage{colortbl}
\usepackage{booktabs}
\usepackage[table]{xcolor}
\usepackage{multirow}
\definecolor{lightblue}{HTML}{EEF6FF}
\newcommand{\spc}{\hspace{1.5pt}}
\renewcommand\footnotetextcopyrightpermission[1]{}
\setcopyright{none}
\acmConference[]{}{}{}
\acmISBN{}
\acmDOI{}
\acmPrice{}

\makeatletter
\renewcommand\@formatdoi[1]{}
\def\@acmConference{}
\def\@acmShortConference{}
\def\@copyrightpermission{}     
\def\@copyrightowner{}           
\def\@acmBooktitle{}             
\makeatother

\begin{document}

\newcommand{\model}{FEP-Diff}
\title{Agent-Centric Social Trajectory Prediction: A Free Energy Principle Perspective}

\author{Yanping Wu}
\affiliation{
  \institution{University of Glasgow}
  \city{Glasgow}
  \country{UK}}
\email{3066431W@student.gla.ac.uk}

\author{Ji Zhang}
\affiliation{
  \institution{Southwest Jiaotong University}
  \city{Chengdu}
  \country{China}}
\email{jizhang.jim@gmail.com}

\author{Hao Chen}
\affiliation{
  \institution{University of Glasgow}
  \city{Glasgow}
  \country{Uk}}
\email{Hao.Chen.2@glasgow.ac.uk}

\author{Edmond S.L. Ho}
\affiliation{
  \institution{University of Glasgow}
  \city{Glasgow}
  \country{UK}}
\email{Shu-Lim.Ho@glasgow.ac.uk}

\author{Chongfeng Wei}
\authornote{Chongfeng Wei is the corresponding author.}
\affiliation{
  \institution{University of Glasgow}
  \city{Glasgow}
  \country{UK}}
\email{Chongfeng.Wei@glasgow.ac.uk}

\renewcommand{\shortauthors}{Yanping Wu et al.}

\begin{abstract}
Trajectory prediction methods have demonstrated remarkable capabilities in capturing complex motion patterns. 
However, existing methods rely on global state assumptions, suffer from insufficient belief inference under partial observability, and lack cognitive behavioral constraints in prediction.
These limitations severely compromise both deployment feasibility and physical plausibility in real-world settings.
In this work, we propose \textit{\model}, an agent-centric trajectory prediction framework grounded in the Free Energy 
Principle, aimed at achieving cognitively plausible predictions under realistic constraints. Specifically, a dual-branch spatiotemporal encoder extracts ego-motion dynamics and social interaction cues from local 
observations. Building upon this, a goal-conditioned belief learner infers multimodal latent belief distributions optimized via a free-energy objective, with a social consistency constraint on the local neighborhood graph to promote cognitive alignment among neighboring agents.
Finally, a residual diffusion trajectory generator is conditioned on the learned belief representations with 
token-level proxy conditioning, producing precise and diverse future predictions. Extensive experiments on five public benchmarks demonstrate that \model\ consistently outperforms state-of-the-art methods under restricted observability. 
Code: \url{https://anonymous.4open.science/r/FEP-Diff-8876}.
\end{abstract}



\received{20 February 2007}
\received[revised]{12 March 2009}
\received[accepted]{5 June 2009}

\maketitle

\section{Introduction}
Trajectory prediction is crucial for agent behavior understanding and motion planning, with broad applications spanning autonomous driving~\cite{10468619,10528911}, mobile robotics~\cite{song2024robot,cheng2020towards}, and pedestrian behavior analysis~\cite{bae2024singulartrajectory,chen2025socialmoif,bae2024can}. 
Recent advances in generative models, particularly Generative Adversarial Networks (GANs)~\cite{gupta2018social} and diffusion models~\cite{bae2024singulartrajectory}, have demonstrated remarkable capability in capturing intricate motion patterns within multi-agent systems (MAS)~\cite{mao2023leapfrog,dendorfer2021mg}.

\begin{figure}[t]
    \centering
    \includegraphics[width=\columnwidth]{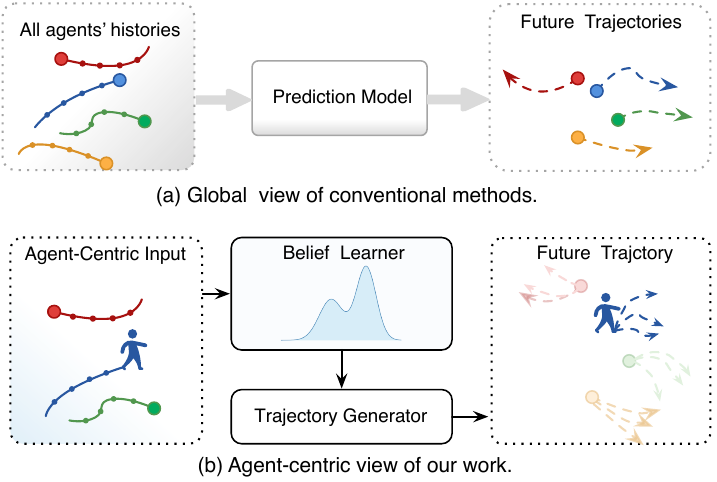}
    \caption{Comparison of trajectory prediction paradigms. 
        (a) Conventional methods assume access to all agents' complete 
        historical trajectories as global input, which is unrealistic 
        in  deployment scenarios. 
        (b) Our \model\ adopts an agent-centric view, where each agent 
        predicts based solely on its local observation. A goal-conditioned belief 
        learner first infers a latent belief distribution grounded in the Free Energy Principle; a residual diffusion 
        trajectory generator then conditions on the frozen belief 
        representations to produce cognitively consistent and 
        physically plausible future trajectories.}
    \label{fig:introd}
\end{figure}

Despite their success, these methods suffer from  three critical limitations.
\textbf{1) Misalignment between global assumptions and real-world deployment:} 
In real-world MAS, agents operate under inherent sensing, communication, and privacy constraints, rendering only partial local observations available for decision-making~\cite{lyu2023risk,ahmed2025distributed,ding2025decentralized}. 
However, existing methods predominantly assume access to a complete global state during both training and inference~\cite{dendorfer2021mg,mao2023leapfrog}---an idealized premise that creates a fundamental misalignment with deployment conditions and severely undermines generalization to real-world scenarios.
\textbf{2) Lack of structured belief inference under partial observability:}
 Existing methods treat trajectory prediction as a purely data-driven regression 
or generation task, without explicitly modeling how agents form and update 
internal beliefs about future states under partial local 
observations~\cite{alahi2016social,salzmann2020trajectron++}.
This passive paradigm lacks the perception--inference loop through which agents 
resolve cognitive discrepancies and achieve social 
consistency~\cite{friston2009free,friston2017active}.
 \textbf{3) Absence of cognitive behavioral constraints:} 
 Despite their powerful multimodal modeling capacity, existing generative paradigms rely solely on statistical distributions to drive the generation 
process, without explicit constraints from high-level behavioral 
principles~\cite{gupta2018social,mao2023leapfrog}.
This renders them prone to generating trajectories that are statistically 
plausible yet cognitively inconsistent with the agent's underlying intent.

These limitations impose a bottleneck on trajectory prediction in decentralized MAS. In real-world scenarios, an ideal prediction paradigm should closely align with the actual decision-making process: agents can only perceive local observations, infer the latent intentions of neighboring agents to achieve behavioral alignment, and generate future trajectories that are grounded in cognitive reasoning.
Although extensive research has been conducted to address individual aspects of this problem (e.g., \cite{omidshafiei2017deep} focuses on decentralized learning under partial observability, while \cite{mao2023leapfrog} emphasizes capturing multimodal distributions),
existing efforts generally lack a unified perspective that jointly captures the intrinsic coupling among local observations, belief inference, and behavioral logic, as illustrated in Figure~\ref{fig:introd}.
This motivates us to revisit trajectory prediction from a new perspective and raises a critical yet underexplored question:
\textit{Can we formulate an agent-centric framework that encodes latent 
beliefs from partial local observations, ensuring that future trajectories 
are underpinned by cognitive behavioral logic?}

To address these limitations, we propose \model, a diffusion-based agent-centric trajectory prediction framework grounded in the Free Energy Principle (FEP), designed to achieve accurate and cognitively consistent predictions in real-world environments.
\textbf{Firstly}, we design a social-aware spatiotemporal encoder that models each agent's local observation in an agent-centric manner, jointly capturing ego-motion dynamics and social interaction cues via a dual-branch architecture with adaptive gated fusion.
\textbf{Secondly}, inspired by the FEP, we introduce a goal-conditioned belief learner that infers multimodal latent belief distributions from local observations, optimized via a free-energy objective with social consistency constraints enforced on the local neighborhood graph.
\textbf{Thirdly}, we propose a residual diffusion trajectory generator that builds upon the frozen belief representations, learning to denoise belief-conditioned trajectory residuals with token-level proxy conditioning for precise and efficient multimodal prediction. We validate \model\ on five standard pedestrian trajectory prediction benchmarks and demonstrate consistent improvements over state-of-the-art 
methods. Notably, our framework maintains a compact architecture, making it well-suited for deployment in real-world settings.
We summarize our contributions below:
\begin{itemize}
    \item We propose \model, an agent-centric trajectory prediction framework grounded in the FEP, which infers structured latent beliefs from partial local observations and enforces social cognitive consistency among neighboring agents.
    \item We introduce a residual diffusion trajectory generator with 
    token-level proxy conditioning that refines belief-conditioned proxy trajectories into precise multimodal predictions via residual denoising.
    \item
    Extensive experiments on five public benchmarks demonstrate that \model\ consistently achieves superior performance over state-of-the-art methods, validating the effectiveness of the proposed framework.
\end{itemize}

\section{Related Work}
\textbf{Trajectory Prediction}.
Although trajectory prediction in MAS has attracted significant attention in recent years~\cite{huang2022survey,singh2023trajectory}, existing paradigms typically operate under idealized deployment assumptions that seldom hold in practice.
Due to inherent limitations such as sensor range, communication bandwidth, and privacy constraints~\cite{maity2021multiagent}, achieving robust trajectory prediction under partial observability presents a fundamentally greater challenge.
Despite their diversity, existing methods—including graph-based~\cite{bae2022learning,mohamed2020social,salzmann2020trajectron++}, interaction-based~\cite{xu2022socialvae,gupta2018social}, and generative models~\cite{mao2023leapfrog,gu2022stochastic,bae2024singulartrajectory}—rely on the unrealistic assumption of global access to all agents' historical trajectories, which fundamentally contradicts the realities of practical deployment. The inability to satisfy this global-access requirement leads to a sharp performance degradation in realistic deployment.
In this work, we provide an agent-centric perspective that eliminates reliance on global states by design, thereby improving both theoretical consistency and real-world applicability.

\textbf{Diffusion models for trajectory prediction}. Generative para\-digms, particularly diffusion models, have demonstrated exceptional efficacy in capturing the complex motion patterns inherent in trajectory prediction~\cite{capellera2025unified,bahari2025certified,bae2024singulartrajectory}.
For example, Bae et al.~\cite{bae2024singulartrajectory} leverage diffusion models to unify human motion patterns across diverse tasks.
However, existing diffusion models predominantly emphasize statistical distribution fitting~\cite{mao2023leapfrog, dendorfer2021mg}, failing to account for the information constraints inherent to actual deployment scenarios.
Despite efforts to embed attention or spatial graph convolutions into diffusion models for capturing interactions~\cite{gu2022stochastic,yang2025trajdiff,mao2023leapfrog},
they typically operate under global-scene assumptions rather than local observation ranges. In contrast to prior works that rely on naive masking strategies to handle data incompleteness~\cite{yang2025unified}, our \model\ conditions the diffusion process on the latent beliefs learned under an FEP-based objective, ensuring cognitive consistency and physical plausibility of generated trajectories under real-world settings.
 
\textbf{Free Energy Principle}.
The FEP provides a principled framework for understanding how agents perform inference and decision-making under uncertainty through active inference~\cite{friston2010free}.
Within the domain of robotics and autonomous systems, prior studies have effectively integrated FEP into state estimation \cite{zhang2024overview,shafiei2025distributionally} and trajectory planning~\cite{mazzaglia2022free,pezzulo2024active}, demonstrating its advantages in maintaining robust internal beliefs despite the presence of observational noise.
Recent FEP-based multi-agent studies~\cite{friston2015knowing} employ Theory of Mind to infer intentions, yet they remain limited to explicit dynamics models and low-dimensional tasks.
In this work, we integrate the FEP into the belief learning stage of a diffusion-based prediction framework. This enables belief-driven inference to compensate for incomplete observations, thereby ensuring cognitive consistency and physical plausibility for trajectory prediction.

\begin{figure*}[htbp]
    \centering
    \includegraphics[height=6.5cm,width=\textwidth]{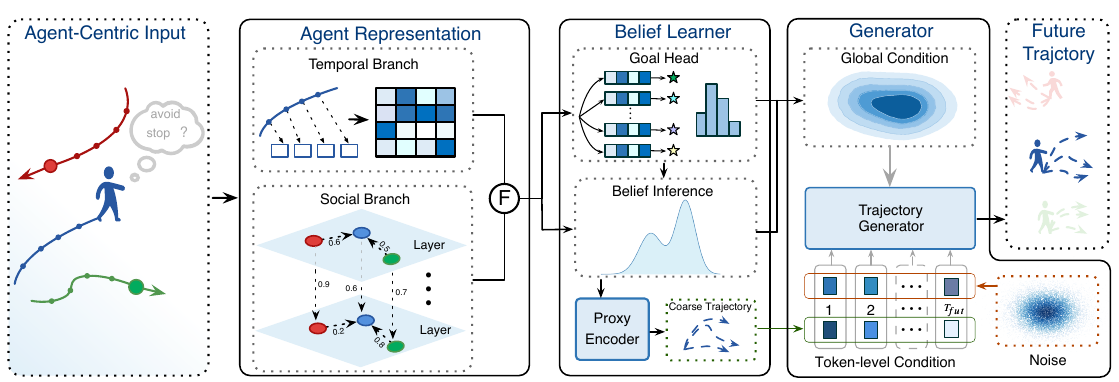}
    \caption{Overview of \model, comprising three 
    modules: an \textit{Agent Representation} encoder that extracts 
    ego-motion and social interaction features from local observations; 
    a \textit{Belief Learner} that infers 
    goal-conditioned latent beliefs from a FEP perspective; 
    and a \textit{Generator} that refines belief-conditioned proxy trajectories into precise multimodal predictions via residual denoising.}
    \label{fig:framework}
\end{figure*}

\section{Preliminaries}
\subsection{Free Energy Principle}
\label{sec:fep}
The FEP provides a theoretical framework for understanding how agents perform inference and decision-making in uncertain environments~\cite{friston2010free,friston2009free}.
It assumes that an agent maintains internal beliefs about the hidden states of the environment and continuously updates these beliefs to reduce uncertainty~\cite{friston2015knowing}. 
Formally, agent behavior can be interpreted as attempting to minimize the \textit{surprise} of its sensory observations.
Technically, \textit{surprise} is defined as the negative log-likelihood of an observation $\mathcal{\boldsymbol{o}}$, denoted as $-\log p(\mathcal{\boldsymbol{o}})$.
Since computing the marginal likelihood $p(\mathcal{\boldsymbol{o}})$ is generally intractable, FEP introduces an approximate posterior (referred to as the \textit{belief distribution})
$q(\boldsymbol{z} \mid \mathcal{\boldsymbol{o}})$ over the latent states $\boldsymbol{z}$.
The agent then minimizes the variational free energy $\mathcal{F}$, which serves as a tractable upper bound on surprise and can be decomposed into:
\begin{equation}
\mathcal{F}=
\underbrace{\mathcal{D}_\mathrm{KL}\!\left[q(\boldsymbol{z} \mid \boldsymbol{o})\,\|\,p(\boldsymbol{z})\right]}_{\mathtt{Complexity}}
+
\underbrace{\mathbb{E}_{q(\boldsymbol{z} \mid \boldsymbol{o})}\! \left[-\ln p(\mathcal{\boldsymbol{o}} \mid \boldsymbol{z})\right]}_{\mathtt{Accuracy}},
\end{equation}
where the $\mathtt{Complexity}$ term penalizes beliefs that deviate significantly from the prior, and the $\mathtt{Accuracy}$ term measures the expected reconstruction error of the local observations under the belief distribution.
Inspired by FEP, we model each agent's internal belief as a goal-conditioned latent 
distribution that encodes future behavioral intent and local interaction context. 
Specifically, the model learns an agent-centric belief distribution by minimizing a 
free-energy-inspired objective (see Fig.~\ref{fig:framework}), providing a principled 
objective for learning structured latent beliefs and encouraging the predicted 
trajectories to reflect cognitively plausible behavioral intent.

\subsection{Diffusion Models}
\label{sec:diff}
Diffusion models have emerged as a highly effective generative paradigm, whose key advantage lies in their ability to model complex multimodal distributions~\cite{ho2020denoising,song2020score,zhu2023difftraj}.
This property makes them particularly well-suited for trajectory generation tasks, where multiple plausible future trajectories may coexist for the same scenario.

\textbf{Forward Process.} The forward diffusion process  gradually perturbs a clean sample $\boldsymbol{x}_0$ with Gaussian noise over $T$ diffusion steps, eventually transforming it into approximately isotropic Gaussian noise. 
At each step, the transition is defined as:
\begin{equation}
q(\boldsymbol{x}_t \mid \boldsymbol{x}_{t-1}) =
\mathcal{N}\!\left(\boldsymbol{x}_t;\sqrt{1-\beta_t}\,\boldsymbol{x}_{t-1},\,\beta_t \boldsymbol{I}\right),
\end{equation}
where $\beta_t \in (0,1)$ denotes the variance schedule at diffusion step $t$. Note that $q$ here denotes the fixed forward Markov kernel of the diffusion process, which is distinct from the learnable variational posterior $q(\boldsymbol{z} \mid \mathcal{\boldsymbol{o}})$ introduced in Sec.~\ref{sec:fep}.
By defining $\alpha_t = 1-\beta_t$ and the cumulative product $\bar{\alpha}_t = \prod_{s=1}^{t}\alpha_s$, one can directly sample $\boldsymbol{x}_t$ from the clean data $\boldsymbol{x}_0$ in closed form:
\begin{equation}
q(\boldsymbol{x}_t \mid \boldsymbol{x}_0) =
\mathcal{N}\!\left(\boldsymbol{x}_t; \sqrt{\bar{\alpha}_t}\,\boldsymbol{x}_0,(1-\bar{\alpha}_t)\boldsymbol{I}
\right).
\label{eq:forward_diff1}
\end{equation}

\textbf{Reverse Process.} The reverse denoising process starts from pure noise $\boldsymbol{x}_T \sim \mathcal{N}(\boldsymbol{0}, \boldsymbol{I})$ and progressively removes noise through a learned denoising network, ultimately recovering a realistic data sample. Each reverse transition is:
\begin{equation}
p_{\theta}(\boldsymbol{x}_{t-1} \mid \boldsymbol{x}_{t}) =
\mathcal{N}\!\left(\boldsymbol{x}_{t-1};\boldsymbol{\mu}_{\theta}(\boldsymbol{x}_{t},t),
\sigma_t^2 \boldsymbol{I}\right),
\end{equation}
where $\boldsymbol{\mu}_{\theta}$ is parameterized by the denoising network.

\textbf{Training and Inference.}
The training objective is typically simplified to predicting the noise $\boldsymbol{\epsilon}$ 
injected during the forward process. The loss function is:
\begin{equation}
\mathcal{L} =
\mathbb{E}_{\boldsymbol{x}_0,\boldsymbol{\epsilon},t}
\left[ \left\|\boldsymbol{\epsilon} - \boldsymbol{\epsilon}_\theta(\boldsymbol{x}_t,t)\right\|^2 \right],
\end{equation}
where $\boldsymbol{\epsilon} \sim \mathcal{N}(\boldsymbol{0}, \boldsymbol{I})$ 
and $t$ is uniformly sampled from $\{1, \dots, T\}$.
At inference time, we adopt Denoising Diffusion Implicit Models 
(DDIM)~\cite{song2020denoising}, which enables deterministic sampling in far fewer steps by skipping intermediate time steps, without retraining the noise prediction network.

\section{Methodology}
we first provide an overview of the \model\ framework along with the problem formulation. We then describe the \textit{Belief Learner}, which infers goal-conditioned latent  belief representations from local observations, followed by the \textit{Trajectory Generator}, 
which employs residual diffusion to produce multimodal future trajectory predictions 
consistent with the inferred beliefs.

\subsection{Framework Overview}
Figure~\ref{fig:framework} provides an overview of the \model\ architecture. Given the agent-centric local observation, the \textit{Agent Representation} module extracts a unified context embedding via a dual-branch encoder, where a Transformer temporal branch captures ego-motion dynamics and a stacked GAT social branch aggregates neighborhood interaction cues, fused through a gated mechanism.
The \textit{Interaction-Driven Belief Learner} decodes the context embedding into candidate goal hypotheses and estimates a goal-conditioned latent belief posterior for each, optimized under a free-energy objective. A lightweight proxy decoder generates coarse trajectory estimates to supervise hypothesis selection during training.
The \textit{Belief-Guided Diffusion Trajectory Generator} conditions on the frozen belief representations, injecting the proxy trajectory as token-level guidance into a residual diffusion model that iteratively denoises trajectory residuals from Gaussian noise into precise and diverse future predictions.

\subsection{Problem Definition}
Trajectory prediction aims to forecast the future motion of agents based on their historical observations and local interaction context. In this work, we consider an \textbf{agent-centric} setting, where each agent $i$ makes predictions based solely on its own \textit{local observation}, rather than relying on global information of the entire scene.
Formally, for a target agent $i$, we define its historical trajectory over $T_\mathtt{obs}$ time steps as $\boldsymbol{X}_i  = \left\{ \boldsymbol{p}_i^t  \mid t = 
-T_{\mathtt{obs}}+1, \dots,0 \right\}$, 
where $\boldsymbol{p}_i^t \in \mathbb{R}^2$ denotes its 2D spatial position at time $t$, and $t=0$ represents the current time step.
Correspondingly, agent $i$ perceives the historical trajectories of surrounding agents within a spatial threshold $\delta$:
\begin{equation}
    \mathcal{S}_i = \{ \boldsymbol{X}_j \mid \|\boldsymbol{p}_j^0 - \boldsymbol{p}_i^0\|_2 < \delta, \; j \in \mathcal{V} \setminus \{i\} \},
\end{equation}
where $\mathcal{V}$ denotes the set of all agents in the scene. 
We then define the local observation of target agent $i$ as:
\begin{equation}
    \mathcal{O}_i = (\boldsymbol{X}_i, \mathcal{S}_i),
    \label{eq:neig}
\end{equation}
which encapsulates both ego-motion history and social interaction cues.
In this paper, our objective is to predict the future trajectory of agent $i$ over the next $T_\mathtt{fut}$ steps, denoted as $\widehat{\boldsymbol{Y}}_i = \{ \widehat{\boldsymbol{p}}_i^t \mid t = 1, \dots, T_\mathtt{fut} \}$. 
This task is formulated as learning the conditional distribution $p\left(\widehat{\boldsymbol{Y}}_i \mid \mathcal{O}_i \right)$.
To handle the inherent uncertainty and multimodality of human behavior, 
the model produces $K$ candidate future trajectories $\{\widehat{\boldsymbol{Y}}_i^{(k)}\}_{k=1}^K$
for each agent $i$.

\subsection{Interaction-Driven Belief Learner}
\subsubsection{\textbf{Social-Aware Spatiotemporal Agent Representation}}
The future behavior of an agent is
jointly shaped by its own motion history and the interactions with surrounding agents. To capture both factors, we design a dual-branch encoder that separately models temporal motion dynamics and spatial social interactions, then fuses them into a unified context representation.

\textbf{Temporal Branch.}
For the temporal branch, we extract kinematic features via discrete 
finite differences, forming $\boldsymbol{m}_i = (\boldsymbol{p}_i, \boldsymbol{v}_i, 
\boldsymbol{a}_i) \in \mathbb{R}^{T_{\mathtt{obs}} \times 6}$, which encodes 
position, velocity, and acceleration at each time step.
The sequence $\boldsymbol{m}_i$ is then fed into a Transformer-based temporal 
encoder $f_{\tau}(\cdot)$ to extract latent motion intent:

\begin{equation}
\boldsymbol{h}_i^\tau = f_{\tau}(\boldsymbol{m}_i), \quad \boldsymbol{h}_i^\tau \in \mathbb{R}^d,
\end{equation}
where $f_\tau(\cdot)$ consists of multiple layers of self-attention~\cite{vaswani2017attention}, with the core operation defined as:
\begin{equation}
\mathtt{SA}(\boldsymbol{m}_i) = \mathtt{softmax} \left( \frac{(\boldsymbol{m}_i \boldsymbol{W}^Q)(\boldsymbol{m}_i \boldsymbol{W}^K)^\top}{\sqrt{d_k}} \right) (\boldsymbol{m}_i \boldsymbol{W}^V),
\end{equation}
where $\boldsymbol{W}^Q$, $\boldsymbol{W}^K$ and $\boldsymbol{W}^V$ are learnable projection matrices and $d_k$ is the key dimension.

\textbf{Social Branch.}
Meanwhile, to model inter-agent interactions, we construct a local interaction graph 
$\mathcal{G}_i = (\mathcal{V}_i, \mathcal{E}_i)$, where the node set 
$\mathcal{V}_i = \{i\} \cup \{j \mid \boldsymbol{X}_j \in \mathcal{S}_i\}$ 
contains the target agent and its surrounding agents, and the edge set $\mathcal{E}_i$ connects pairs of agents within the spatial threshold $\delta$.
A stacked Graph Attention Network (GAT)~\cite{velivckovic2017graph} is applied over $\mathcal{G}_i$ 
to aggregate neighborhood interaction cues.
For each edge $(i,j) \in \mathcal{E}_i$, we construct a 6-dimensional 
rotation-invariant edge feature $\boldsymbol{e}_{ij}$ encoding the relative 
geometric and motion relationship between agents $i$ and $j$.
The attention weight is computed as:
\begin{equation}
\alpha_{ij} = \frac{\exp \left( \psi(\boldsymbol{W}\boldsymbol{h}_i^\tau,\, 
\boldsymbol{W}\boldsymbol{h}_j^\tau,\, \boldsymbol{e}_{ij}) \right)}
{\sum_{k \in \mathcal{V}_i} \exp \left( \psi(\boldsymbol{W}\boldsymbol{h}_i^\tau,\, 
\boldsymbol{W}\boldsymbol{h}_k^\tau,\, \boldsymbol{e}_{ik}) \right)},
\end{equation}
where $\psi(\cdot)$ is a scoring function implemented as a linear layer 
followed by LeakyReLU activation, and $\boldsymbol{W}$ denotes the node feature projection matrix.
The interaction-aware representation of agent $i$ is then obtained as:
\begin{equation}
\boldsymbol{h}_i^{\mathtt{soc}} = \sum_{j \in \mathcal{V}_i} \alpha_{ij} 
\left(\boldsymbol{W}\boldsymbol{h}_j^\tau + \boldsymbol{W}_e\boldsymbol{e}_{ij}\right), 
\quad \boldsymbol{h}_i^{\mathtt{soc}} \in \mathbb{R}^d.
\end{equation}
By stacking multiple GAT layers, the model captures higher-order interaction 
dependencies within the local neighborhood.

\textbf{Gated Fusion.}
Since the future motion of an agent is governed by both its intrinsic tendency and external social pressure, we introduce a gated fusion mechanism to combine the two branches:
\begin{equation}
    \boldsymbol{g}_i = \boldsymbol{W}_g [\boldsymbol{h}_i^\tau \parallel \boldsymbol{h}_i^{\mathtt{soc}}] + \boldsymbol{b}_g,
\end{equation}
where $\boldsymbol{g}_i$ serves as a soft gate that dynamically weights the contribution of each branch, and the fused representation is:
\begin{equation}
    \boldsymbol{h}_i = \sigma(\boldsymbol{g}_i) \odot \phi^{\mathtt{self}}(\boldsymbol{h}_i^\tau) + (1 - \sigma(\boldsymbol{g}_i)) \odot \phi^{\mathtt{soc}}(\boldsymbol{h}_i^{\mathtt{soc}}),
\end{equation}
where $\sigma(\cdot)$ denotes the Sigmoid activation function, $\odot$ denotes element-wise multiplication, and $\phi^{\mathtt{self}}(\cdot)$, $\phi^{\mathtt{soc}}(\cdot)$ are learnable projection functions.
 Consequently, $\boldsymbol{h}_i \in \mathbb{R}^d$ jointly encodes the individual motion and social context of agent $i$, serving as the context embedding for subsequent belief inference.

\subsubsection{\textbf{Goal-Driven Belief Posterior}}
To handle the inherent multimodality of agent future behavior, the context embedding $\boldsymbol{h}_i$  is decoded into $K$ discrete goal hypotheses and their corresponding importance weights:
\begin{equation}
\{\boldsymbol{p}_k^*\}_{k=1}^K = f_{\mathtt{goal}}(\boldsymbol{h}_i), 
\quad \{\pi_k\}_{k=1}^K = \mathtt{softmax}(f_{w}(\boldsymbol{h}_i)),
\end{equation}
where $\boldsymbol{p}_k^* \in \mathbb{R}^2$ denotes the $k$-th candidate estimate 
of the future endpoint $\boldsymbol{p}_i^{T_{\mathtt{fut}}}$, and $\pi_k$ 
represents its normalized likelihood.
Conditioned on each goal hypothesis $\boldsymbol{p}_k^* $, a goal-conditioned 
belief head estimates the parameters of a Gaussian posterior over the latent 
state $\boldsymbol{z}_i \in \mathbb{R}^{d_z}$:

\begin{equation}
(\boldsymbol{\mu}_{z_k},\, \log \boldsymbol{\sigma}_{z_k})
= f_{\mathtt{belief}}(\boldsymbol{h}_i,\, \boldsymbol{p}_k^*),
\end{equation}
which defines the Gaussian posterior:
\begin{equation}
q(\boldsymbol{z}_i \mid \mathcal{O}_i,\, \boldsymbol{p}_k^*)
= \mathcal{N}\!\left(\boldsymbol{\mu}_{z_k},\, 
\mathrm{diag}(\boldsymbol{\sigma}_{z_k}^2)\right).
\end{equation}
This corresponds to the perceptual inference process under the FEP 
(Sec.~\ref{sec:fep}), where the agent updates its internal belief conditioned 
on local observations and hypothetical goals.

To preserve the supervisory signal for the belief learner, we employ a 
lightweight proxy decoder to generate an initial trajectory estimate for 
each hypothesis:
\begin{equation}
\widetilde{\boldsymbol{Y}}_i^{(k)} = f_{\mathtt{proxy}}\!\left(
\boldsymbol{\mu}_{z_k},\, \boldsymbol{p}_k^*\right),
\end{equation}
where $\widetilde{\boldsymbol{Y}}_i^{(k)} = \{\widetilde{\boldsymbol{p}}_{i,t}^{(k)} 
\mid t = 1, \dots, T_{\mathtt{fut}}\}$ denotes the proxy trajectory for 
hypothesis $k$. Here, only the mean $\boldsymbol{\mu}_{z_k}$ is passed to the 
decoder as a deterministic belief representation, avoiding stochastic sampling 
during training.

\subsubsection{\textbf{Individual Free Energy}}
Given the goal-conditioned belief posterior, we formulate the learning objective based on the FEP. For each agent $i$ and hypothesis $k$, the free energy is defined as:
\begin{equation}
\begin{aligned}
\mathcal{F}_i^{(k)} = & \underbrace{\| \widetilde{\boldsymbol{Y}}_{i}^{(k)} - 
\boldsymbol{Y}_i \|^2}_{\text{Accuracy}} \\
+ & \lambda_{\mathrm{kl}} \underbrace{\sum_{d=1}^{d_z} \max \left( \mathcal{D}_{\mathrm{KL}}^{(d)} 
\left[ q(\boldsymbol{z}_i \mid \mathcal{O}_i, \boldsymbol{p}_k^*) \| 
p(\boldsymbol{z}) \right], \lambda_{\mathtt{fb}} \right)}_{\text{Complexity}},
\label{eq:ind_fe}
\end{aligned}
\end{equation}
where $\boldsymbol{Y}_i$ is the ground-truth future trajectory of agent $i$, 
the $\mathtt{Accuracy}$ term measures the reconstruction error between the proxy 
prediction and ground truth, and the $\mathtt{Complexity}$ term penalizes the 
KL divergence of the $d$-th latent dimension between the posterior 
$q(\boldsymbol{z}_i \mid \mathcal{O}_i, \boldsymbol{p}_k^*)$ and the prior 
$p(\boldsymbol{z})$~\cite{kingma2013auto}. A free-bits strategy lower-bounds each dimension 
by $\lambda_{\mathtt{fb}}$ to prevent posterior collapse~\cite{kingma2016improved}.

However, optimizing all hypotheses simultaneously may lead to mode collapse and unstable training. To address this, we adopt a winner-take-all (WTA) strategy that selects the best-matching hypothesis by endpoint proximity~\cite{zhang2022distributed}:
\begin{equation}
\label{eq:wta}
k_i^* = \arg\min_k \|\boldsymbol{p}_k^* - \boldsymbol{p}_i^{T_{\mathtt{fut}}}\|_2,
\end{equation}
where $\boldsymbol{p}_i^{T_{\mathtt{fut}}}$ denotes the ground-truth endpoint. 
The individual free energy is then computed on the selected hypothesis only:
\begin{equation}
\mathcal{F}_i = \mathcal{F}_i^{(k_i^*)}.
\end{equation}

To further stabilize training and preserve multimodality, we introduce a 
diversity constraint and a classification loss over all $K$ hypotheses:
\begin{equation}
\mathcal{L}_{\mathtt{div}} = \frac{1}{K(K-1)} \sum_{k \neq k'} 
\left[ \mathtt{ReLU}\!\left(m - \|\boldsymbol{p}_k^* - 
\boldsymbol{p}_{k'}^*\|_2\right) \right]^2,
\end{equation}
\begin{equation}
\mathcal{L}_{\mathtt{cls}} = \mathtt{CE}(\boldsymbol{\pi}, k_i^*),
\label{eq:goal_sup}
\end{equation}
where $m$ is a predefined margin enforcing minimum spatial separation among 
candidate endpoints, and $\boldsymbol{\pi} = \{\pi_k\}_{k=1}^K$ is the 
predicted categorical distribution over hypotheses.

\subsubsection{\textbf{Social Free Energy}}
Intuitively, agents within the same local neighborhood are exposed to highly correlated environmental observations, and thus their variational free energy tends to remain consistent. 
Motivated by this, we introduce a social free energy consistency constraint to capture the coherent perceptual inference process among neighboring agents. 
Specifically, for each pair $(i,j) \in \mathcal{E}$, where 
$\mathcal{E} = \bigcup_i \mathcal{E}_i$ denotes the set of all neighboring 
agent pairs, we evaluate each 
agent's posterior under its respective best-matching hypothesis 
$k_i^*$ and $k_j^*$ (Eq.~\ref{eq:wta}), and penalize discrepancies 
using a symmetric KL divergence:
\begin{equation}
\begin{aligned}
\mathcal{L}_{\mathtt{cons}} 
&= \frac{1}{2|\mathcal{E}|}\sum_{(i,j)\in\mathcal{E}} 
\left[
\mathcal{D}_{\mathrm{KL}}\bigl(q(\boldsymbol{z}_i \mid \mathcal{O}_i, \boldsymbol{p}_{k_i^*}^*)\,\|\,
q(\boldsymbol{z}_j \mid \mathcal{O}_j, \boldsymbol{p}_{k_j^*}^*)\bigr) \right.\\
&\left.\quad+ 
\mathcal{D}_{\mathrm{KL}}\bigl(q(\boldsymbol{z}_j \mid \mathcal{O}_j, \boldsymbol{p}_{k_j^*}^*)\,\|\,
q(\boldsymbol{z}_i \mid \mathcal{O}_i, \boldsymbol{p}_{k_i^*}^*)\bigr)
\right].
\end{aligned}
\label{eq:social_cons}
\end{equation}
We further adopt a collision-avoidance constraint in proxy trajectory space:
\begin{equation}
\mathcal{L}_{\mathtt{coll}} 
= \frac{1}{|\mathcal{E}|}\sum_{(i,j)\in\mathcal{E}} 
\left[
\mathtt{ReLU}\!\left(d_{\min} - \min_t\|\widetilde{\boldsymbol{p}}_{i,t}^{(k_i^*)} - 
\widetilde{\boldsymbol{p}}_{j,t}^{(k_j^*)}\|_2\right)
\right]^2,
\end{equation}
where $d_{\min}$ is a safety distance threshold.
The social free energy is then defined as:
\begin{equation}
\mathcal{F}_{\mathtt{social}} 
= \lambda_{\mathtt{cons}}\mathcal{L}_{\mathtt{cons}} 
+ \lambda_{\mathtt{coll}}\mathcal{L}_{\mathtt{coll}},
\label{eq:social}
\end{equation}

which enforces both cognitive consistency in the latent space and physical 
feasibility in the trajectory space.
The overall training objective is formulated as:
\begin{equation}
\mathcal{L}_{\mathtt{total}}
=
\frac{1}{|\mathcal{V}|}\sum_i \mathcal{F}_{i}
+ \lambda_{\mathtt{cls}}\mathcal{L}_{\mathtt{cls}}
+ \lambda_{\mathtt{div}}\mathcal{L}_{\mathtt{div}}
+ \mathcal{F}_{\mathtt{social}}.
\end{equation}

\subsection{Belief-Guided Diffusion Trajectory Generator}
To address the capacity bottleneck of the proxy decoder, we introduce a 
residual diffusion model that refines the proxy trajectory into precise 
future predictions by learning to denoise belief-conditioned trajectory 
residuals.

\subsubsection{\textbf{Trajectory-aware Conditioning}}
For each hypothesis $k$, we first construct a global condition vector by concatenating the latent belief mean and the goal endpoint:
\begin{equation}
\boldsymbol{c}_k = [\,\boldsymbol{\mu}_{z_k};\, \boldsymbol{p}_k^*\,] 
\in \mathbb{R}^{d_z+2}.
\end{equation}
Unlike standard conditional diffusion models that treat the predicted trajectory as an unstructured vector, we explicitly inject the proxy trajectory as a \emph{token-level condition}—a design we refer to as 
\emph{trajectory-aware denoising}. Specifically, the proxy trajectory is:
\begin{equation}
\widetilde{\boldsymbol{Y}}^{(k)}_{i} = 
\left\{\widetilde{\boldsymbol{p}}^{(k)}_{i,t} \mid t = 1,\dots,T_{\mathtt{fut}}\right\},
\end{equation}
where $\widetilde{\boldsymbol{p}}^{(k)}_{i,t} \in \mathbb{R}^2$ is the proxy 
position at future time step $t$ under hypothesis $k$.

Instead of directly generating the final trajectory, the diffusion model learns 
a residual:
\begin{equation}
\Delta\boldsymbol{Y}^{(k)}_i = \boldsymbol{Y}_i - \widetilde{\boldsymbol{Y}}^{(k)}_i,
\end{equation}
so that the final prediction is recovered as:
\begin{equation}
\widehat{\boldsymbol{Y}}^{(k)}_i = \widetilde{\boldsymbol{Y}}^{(k)}_i + 
\Delta\boldsymbol{Y}^{(k)}_i.
\label{eq:final_refine}
\end{equation}
At each denoising step, the model is conditioned not only on the current noisy 
residual but also on the proxy trajectory state at every future time step, 
enabling step-wise refinement with explicit temporal and geometric awareness.

\subsubsection{\textbf{Residual Diffusion Formulation}}
To stabilize training, the residual target is normalized using dataset-level statistics:
\begin{equation}
\boldsymbol{x}_0 = \frac{\Delta\boldsymbol{Y}^{(k)}_i - \boldsymbol{\mu}_{r}}
{\boldsymbol{\sigma}_{r}},
\end{equation}
where $\boldsymbol{\mu}_{r}$ and $\boldsymbol{\sigma}_{r}$ are the mean and 
standard deviation of training residuals.
The forward diffusion process follows Eq.~\ref{eq:forward_diff1}, adding noise 
to $\boldsymbol{x}_0$ over $T$ steps.
The denoiser $\mathcal{D}_{\theta}$ takes as input the noisy residual 
$\boldsymbol{x}_t$, diffusion timestep $t$, global condition $\boldsymbol{c}_k$, 
and the proxy trajectory $\widetilde{\boldsymbol{Y}}^{(k)}_{i}$, which is formulated as:
\begin{equation}
\hat{\boldsymbol{\epsilon}} =
\mathcal{D}_{\theta}\!\left(\boldsymbol{x}_t,\,t,\,
\boldsymbol{c}_k,\,
\widetilde{\boldsymbol{Y}}^{(k)}_{i}
\right).
\label{eq:eps_pred}
\end{equation}
To realize trajectory-aware denoising, the proxy trajectory is injected at 
the token level. For each future time step $t' \in \{1,\dots,T_{\mathtt{fut}}\}$, 
the input token is formed as:
\begin{equation}
\boldsymbol{u}_{t'}
=
\left[
\boldsymbol{r}_{t'};\,
\widetilde{\boldsymbol{p}}^{(k)}_{i,t'}
\right]
\in\mathbb{R}^{4},
\label{eq:token_input}
\end{equation}
where $\boldsymbol{r}_{t'} \in \mathbb{R}^2$ is the noisy residual component 
at time step $t'$, and $\widetilde{\boldsymbol{p}}^{(k)}_{i,t'} \in 
\mathbb{R}^2$ is the corresponding proxy trajectory coordinate. This design makes each denoising step explicitly aware of the proxy trajectory state, enabling precise step-wise refinement rather than treating the entire trajectory as a flat unstructured vector.

\subsubsection{\textbf{Training Objective and Inference}}
The \textit{Trajectory Generator} is trained with the standard $\epsilon$-prediction loss 
and Min-SNR weighting~\cite{hang2023efficient}:
\begin{equation}
\mathcal{L}_{\mathtt{diff}}
=
\mathbb{E}_{\boldsymbol{x}_0,t,\boldsymbol{\epsilon}}
\left[
w(t)\,
\left\|\hat{\boldsymbol{\epsilon}} - \boldsymbol{\epsilon}\right\|_2^2
\right],
\label{eq:diff_loss}
\end{equation}
where $w(t) = 1/(\mathtt{SNR}(t)+1)$ and $\mathtt{SNR}(t) = 
\bar{\alpha}_t/(1-\bar{\alpha}_t)$. This weighting suppresses easy timesteps and improves training stability.

At inference time, the \textit{Belief Learner} is frozen and provides all $K$ 
hypotheses in parallel. For each hypothesis $k$, a normalized residual 
is sampled via deterministic DDIM~\cite{song2020denoising}:
\begin{equation}
\hat{\boldsymbol{x}}_0=
\mathtt{DDIM}\!\left(\boldsymbol{c}_k,\,\widetilde{\boldsymbol{Y}}^{(k)}_i\right),
\end{equation}
This sample is then de-normalized to recover the residual:
\begin{equation}
\Delta\boldsymbol{Y}^{(k)}_i =
\hat{\boldsymbol{x}}_0 \odot \boldsymbol{\sigma}_r + \boldsymbol{\mu}_r,
\label{eq:denorm_resid}
\end{equation}
where $\odot$ denotes element-wise multiplication.
The final trajectory is obtained via Eq.~(\ref{eq:final_refine}), 
with all $K$ hypotheses processed as a single batched tensor for 
efficient parallel sampling.

\begin{table*}[h]
\centering
\normalsize
\caption{Stochastic multimodal trajectory prediction comparison with state-of-the-art methods on ETH/UCY benchmarks.
Results are reported as minADE$_{20}$/minFDE$_{20}$ (lower is better). \textbf{Bold}: best. \underline{Underlined}: second best.}
\label{tab:stochastic_results}
\setlength{\tabcolsep}{7pt}
\renewcommand{\arraystretch}{0.8}
\begin{tabular}{l  c c c c c c c}
\toprule
Stochastic   & Venue & ETH & HOTEL & UNIV & ZARA1 & ZARA2 & Average\\
\midrule
Trajectron++~\cite{salzmann2020trajectron++}   &ECCV'20  &0.63\spc/\spc1.06  &0.22\spc/\spc0.34  &0.36\spc/\spc0.60 &0.29\spc/\spc0.43  &0.20\spc/\spc0.35 &0.34\spc/\spc0.56\\
AgentFormer~\cite{yuan2021agentformer}        & ICCV'21 & 0.45\spc/\spc0.75 & 0.14\spc/\spc0.22 & 0.25\spc/\spc0.45 & 0.18\spc/\spc\textbf{0.30} & \underline{0.14}\spc/\spc0.24 & 0.23\spc/\spc0.39 \\
GP-Graph~\cite{bae2022learning}               & ECCV'22 & 0.43\spc/\spc0.64 & 0.18\spc/\spc0.30 & 0.24\spc/\spc0.42 &\underline{0.17}\spc/\spc\underline{0.31} & 0.16\spc/\spc0.29 & 0.24\spc/\spc0.39 \\
Groupnet~\cite{xu2022groupnet}     & CVPR'22 &0.46\spc/\spc0.73 &0.15\spc/\spc0.25  &0.26\spc/\spc0.49 &0.21\spc/\spc0.39  &0.17\spc/\spc0.33  &0.25\spc/\spc0.44  \\
NPSN~\cite{bae2022non}          & CVPR'22 &\underline{0.36}\spc/\spc0.59    &0.16\spc/\spc0.25  &\underline{0.23}\spc/\spc\underline{0.39} &0.18\spc/\spc0.32  &\underline{0.14}\spc/\spc0.25  &\underline{0.21}\spc/\spc0.36  \\
MID~\cite{gu2022stochastic}                   &CVPR'22   &0.47\spc/\spc0.90    & 0.18\spc/\spc0.31   &0.24\spc/\spc0.47   &0.22\spc/\spc0.45     &0.15\spc/\spc0.30 &0.25\spc/\spc0.49  \\
SocialVAE~\cite{xu2022socialvae}              &ECCV'22   &0.42\spc/\spc0.59    &0.14\spc/\spc0.19   &\textbf{0.22}\spc/\spc\textbf{0.37}   &0.18\spc/\spc\textbf{0.30}    &0.15\spc/\spc0.28  &0.22\spc/\spc0.35  \\
EigenTrajectory~\cite{bae2023eigentrajectory}   & ICCV'23 &\underline{0.36}\spc/\spc0.57 & \underline{0.13}\spc/\spc0.21 & 0.24\spc/\spc0.43 & 0.20\spc/\spc0.35 & 0.15\spc/\spc0.26 & 0.22\spc/\spc0.36 \\
EqMotion~\cite{xu2023eqmotion}              & CVPR'23 & 0.40\spc/\spc0.61 & \textbf{0.12}\spc/\spc\underline{0.18} & \underline{0.23}\spc/\spc0.43 & 0.18\spc/\spc0.32 & \textbf{0.13}\spc/\spc\underline{0.23} &\underline{0.21}\spc/\spc0.35  \\
SingularTraj~\cite{bae2024singulartrajectory}   &CVPR'24 &\textbf{0.35}\spc/\spc\textbf{0.42} &\underline{0.13}\spc/\spc0.19 &0.25\spc/\spc0.44 &0.19\spc/\spc0.32 &0.15\spc/\spc0.25 &\underline{0.21}\spc/\spc\textbf{0.32}\\
MoFlow~\cite{fu2025moflow}  &CVPR'25   &0.40\spc/\spc0.59    &\textbf{0.12}\spc/\spc\textbf{0.17}   &0.25\spc/\spc0.41   &\underline{0.17}\spc/\spc\textbf{0.30}    &\textbf{0.13}\spc/\spc\textbf{0.22}  &\underline{0.21}\spc/\spc\underline{0.34}  \\
\rowcolor{lightblue}
\textbf{\model\ (ours)} & --   &0.37\spc/\spc\underline{0.51}  &\textbf{0.12}\spc/\spc\underline{0.18} &\textbf{0.22}\spc/\spc\underline{0.39} &\textbf{0.16}\spc/\spc\textbf{0.30} &\underline{0.14}\spc/\spc0.24 &\textbf{0.20}\spc/\spc\textbf{0.32}  \\
\bottomrule
\end{tabular}
\end{table*}

\section{Experiments}
\subsection{Experimental Setup}
\subsubsection{Datasets and Baselines} 
 To validate \model, we conduct experiments on the ETH~\cite{pellegrini2009you} and UCY~\cite{lerner2007crowds} benchmarks, which contain five distinct scenes (ETH, HOTEL, UNIV, ZARA1, and ZARA2) with diverse social interactions. These datasets provide pedestrian trajectories sampled at $0.4$s intervals. 
 Following the common setup of the trajectory prediction task~\cite{bae2022non,shi2021sgcn,gupta2018social},
we adopt the leave-one-out strategy for cross-validation. 
We compare \model\ with a broad spectrum of baseline methods, including graph-based methods (GP-Graph~\cite{bae2022learning}, Groupnet~\cite{xu2022groupnet}, Trajectron++~\cite{salzmann2020trajectron++}), interaction-based models (EqMotion~\cite{xu2023eqmotion}), generative models (MID~\cite{gu2022stochastic}, SingularTraj~\cite{bae2024singulartrajectory}, MoFlow~\cite{fu2025moflow}), and other representative methods (NPSN~\cite{bae2022non}, SocialVAE~\cite{xu2022socialvae}, EigenTrajectory~\cite{bae2023eigentrajectory}, AgentFormer~\cite{yuan2021agentformer}).

\subsubsection{Evaluation Metrics and Protocol}
Following the established literature~\cite{gupta2018social, alahi2016social}, we evaluate \model\ using two standard metrics: Average Displacement Error (ADE) and Final Displacement Error (FDE). 
We predict $12$ future frames ($4.8$s) conditioned on an $8$-frame ($3.2$s)
observation window.
For stochastic predictions, we adopt the Best-of-$K$ ($K=20$) protocol;
for deterministic predictions, a single sample ($K=1$) is used. All results are averaged over three runs.

\subsubsection{Implementation Details} 
All experiments are conducted in PyTorch 2.11.0 on a server with an Intel i9-14900K CPU and an NVIDIA RTX PRO 6000 GPU.
For the \textit{Belief Learner}, the hidden, GAT output, and latent belief dimensions are set to $128$, $64$, and $128$, with $8$ attention heads and $K=20$ goal hypotheses.
It is optimized with Adam (lr=$10^{-3}$) for up to $150$ epochs with early stopping. Loss weights are 
set to $\lambda_{\mathtt{kl}}=5\times10^{-3}$, $\lambda_{\mathtt{cls}}=0.5$, $\lambda_{\mathtt{div}}=0.25$, $\lambda_{\mathtt{fb}}=0.02$, with $\lambda_{\mathtt{cons}}$ and $\lambda_{\mathtt{coll}}$ tuned per dataset.
For the \textit{Trajectory Generator}, we adopt a linear noise schedule with $T=200$ diffusion steps and  DDIM sampling with $50$ steps at inference. It is trained 
with AdamW (lr=$10^{-4}$) using a $5$-epoch linear warmup followed by cosine decay, and EMA with decay $0.999$, for up to $150$ epochs.

\begin{figure*}[htbp]
    \centering
    \includegraphics[height=6.5cm,width=\textwidth]{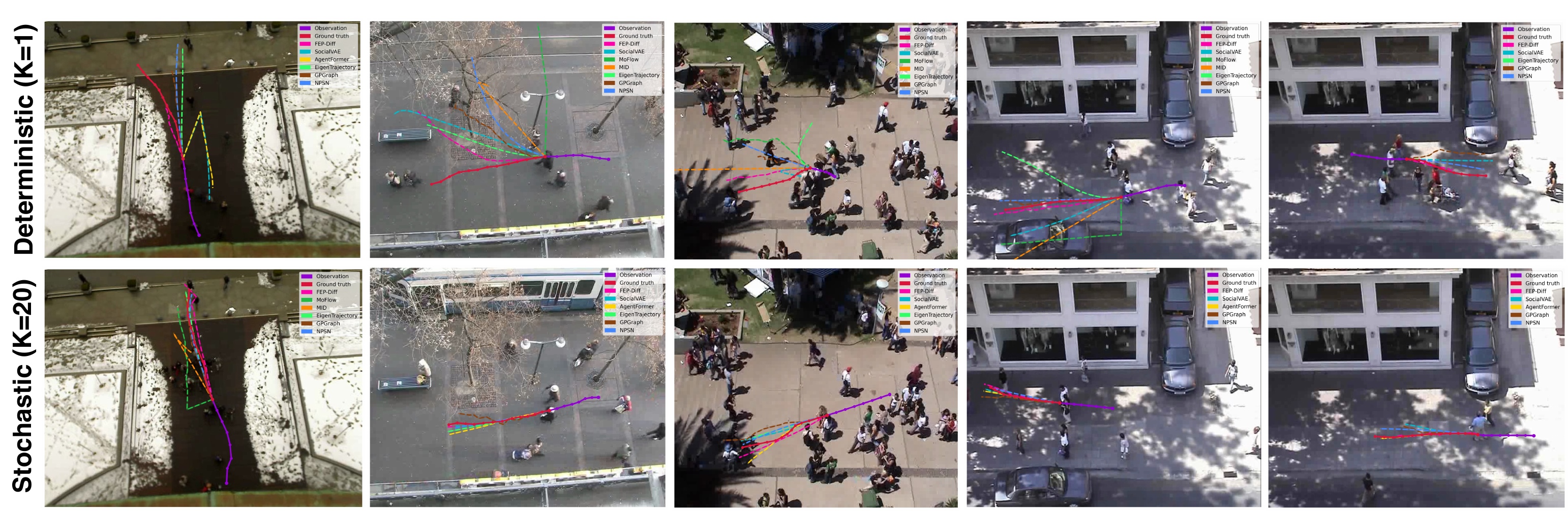}
   \caption{Qualitative comparison of stochastic ($K=20$, bottom) and deterministic ($K=1$, top) trajectory predictions across five benchmarks (ETH, Hotel, UNIV, ZARA1, ZARA2). For the stochastic setting, the best-of-20 sample is shown for each method.}
    \label{fig:qualitative}
\end{figure*}

\subsection{Experimental Results}
\subsubsection{Stochastic Trajectory Prediction Performance}
Table~\ref{tab:stochastic_results} reports stochastic multimodal trajectory
prediction results. 
\model\ achieves the best average performance with minADE$_{20}$/minFDE$_{20}$ of \textbf{0.20/0.32}, outperforming all baselines.
Compared with SingularTraj~\cite{bae2024singulartrajectory}, \model\ achieves
superior average ADE and consistently lower errors on HOTEL, UNIV, and ZARA1;
on ZARA1, ADE improves from 0.19 to \textbf{0.16} with a matched best FDE of
\textbf{0.30}.
Against the recent MoFlow~\cite{fu2025moflow}, \model\ achieves better average performance (\textbf{0.20}/\textbf{0.32} vs. 0.21/0.34), demonstrating the advantage of belief-guided residual diffusion over purely distribution-fitting generative approaches.
The stochastic predictions ($K=20$) are visualized in Figure~\ref{fig:qualitative} (\textit{bottom}).
\model\ generates at least one sample  close to the
ground-truth, confirming effective coverage of plausible future motions.
Notably, all the above results are obtained under the \textbf{agent-centric partial
observability setting}, without access to global scene information assumed by
all compared baselines, highlighting the practical advantage of our framework.

\subsubsection{Deterministic trajectory prediction performance.}
Table~\ref{tab:deterministic} pres\-ents the deterministic trajectory prediction results across five scenes.
Compared with SingularTraj~\cite{bae2024singulartrajectory}, the best-performing baseline, \model\ achieves competitive or superior results on ZARA1 (\textbf{0.39}/\textbf{0.85} vs.\
0.44/0.93) and UNIV (\textbf{0.46}/\textbf{1.00} vs.\ 0.57/1.12), while performing worse on ETH, which we attribute to its larger scene scale, which is less compatible with the local observation assumption.
As shown in Figure~\ref{fig:qualitative} (\textit{top}), predictions remain accurate in the near-term horizon but exhibit increasing deviation at later time steps, attributable to error compounding over time and the inherent difficulty of committing to a single mode under long-horizon uncertainty.
Overall, \model\ achieves an average ADE$_1$/FDE$_1$ of \textbf{0.49/0.95}, ranking second among all compared methods and outperforming the majority of baselines.
Notably, \model\ achieves stronger FDE than ADE rankings across scenes, suggesting that the explicit goal hypothesis mechanism in the \textit{Belief Learner} supervises endpoint prediction and encourages trajectories to converge toward plausible destinations.

\begin{table*}[htbp]
\centering
\normalsize
\caption{Deterministic trajectory prediction comparison with state-of-the-art
methods on ETH/UCY benchmarks. Results are reported as ADE$_1$/FDE$_1$ (lower is better). \textbf{Bold}: best. \underline{Underlined}: second best.}
\label{tab:deterministic}
\renewcommand{\arraystretch}{0.8}
\setlength{\tabcolsep}{7pt}
\begin{tabular}{l c c c c c c}
\toprule
Deterministic & ETH & HOTEL & UNIV & ZARA1 & ZARA2 & AVG \\
\midrule
AgentFormer~\cite{yuan2021agentformer}        &1.03\spc/\spc2.03 &1.41\spc/\spc3.29 &0.67\spc/\spc1.42 &0.82\spc/\spc1.88 &0.70\spc/\spc1.61 &0.93\spc/\spc2.05 \\
EigenTrajectory~\cite{bae2023eigentrajectory} &1.33\spc/\spc2.65 &0.95\spc/\spc1.76 &0.65\spc/\spc1.31 &0.69\spc/\spc1.46 &0.92\spc/\spc1.77 &0.91\spc/\spc1.79 \\
GP-Graph~\cite{bae2022learning}               &0.88\spc/\spc1.78 &0.47\spc/\spc1.03 &0.56\spc/\spc1.19 &\underline{0.40}\spc/\spc0.87 &0.35\spc/\spc0.77 &0.53\spc/\spc1.13 \\
EqMotion~\cite{xu2023eqmotion}              &0.96\spc/\spc1.92 &\underline{0.30}\spc/\spc0.58 &0.50\spc/\spc1.10 &\textbf{0.39}\spc/\spc\underline{0.86} &\textbf{0.30}\spc/\spc\textbf{0.68} &\underline{0.49}\spc/\spc1.03\\
NPSN~\cite{bae2022non}                      &\underline{0.81}\spc/\spc1.67  &0.50\spc/\spc1.03 &0.54\spc/\spc1.14  &0.42\spc/\spc0.92  &\underline{0.32}\spc/\spc\underline{0.69}&0.52\spc/\spc1.09 \\
Trajectron++~\cite{salzmann2020trajectron++}
& 1.06\spc/\spc2.34 &0.38\spc/\spc0.98 &0.56\spc/\spc1.45 &0.48\spc/\spc1.31 &0.35\spc/\spc1.01 &0.57\spc/\spc1.42 \\
SingularTraj~\cite{bae2024singulartrajectory}  &\textbf{0.72}\spc/\spc\textbf{1.23} &\textbf{0.27}\spc/\spc\textbf{0.50} &0.57\spc/\spc1.12 &0.44\spc/\spc0.93 &0.35\spc/\spc0.73  &\textbf{0.47}\spc/\spc\textbf{0.90}\\
MID~\cite{gu2022stochastic} &0.85\spc/\spc1.98 &0.42\spc/\spc0.92 &\textbf{0.43}\spc/\spc\underline{1.01} &0.47\spc/\spc1.10 &0.33\spc/\spc0.78 &0.50\spc/\spc1.16  \\
SocialVAE~\cite{xu2022socialvae}  &0.91\spc/\spc1.80 &0.34\spc/\spc0.71 &0.51\spc/\spc1.11 &0.44\spc/\spc1.01 &0.33\spc/\spc0.75 &0.51\spc/\spc1.08    \\ 
MoFlow~\cite{fu2025moflow}  &1.20\spc/\spc2.60 &0.46\spc/\spc0.98 &0.93\spc/\spc1.77 &0.66\spc/\spc1.42 &1.25\spc/\spc2.72 &0.90\spc/\spc1.90  \\
\rowcolor{lightblue}
\textbf{\model\ (ours)}     &0.94\spc/\spc\underline{1.58} &\underline{0.30}\spc/\spc\underline{0.56} &\underline{0.46}\spc/\spc\textbf{1.00} &\textbf{0.39}\spc/\spc\textbf{0.85} & 0.34\spc/\spc0.77&\underline{0.49}\spc/\spc\underline{0.95} \\
\bottomrule
\end{tabular}%
\end{table*}

\begin{table*}[htbp]
  \setlength{\abovecaptionskip}{0.1cm}
  \setlength{\belowcaptionskip}{-0.2cm}
  \centering
   \caption{Ablation study of different components in \model. ``w/o'' denotes removing the corresponding component. Results are reported as minADE$_{20}$/minFDE$_{20}$ (lower is better). \textbf{Bold}: best, \underline{Underlined}: second best.}
  \label{tab:ablation}
  \setlength{\tabcolsep}{4pt}
  \normalsize
  \renewcommand{\arraystretch}{0.9}
  \begin{tabular}{c|l|cccccc}
  \hline
  Module & Variant & ETH &HOTEL &UNIV &ZARA1 &
ZARA2 &AVG \\
  \hline
\multirow{3}{*}{\shortstack{Belief \\ Learner}}
  & \textcircled{\small 1} w/o Individual Free Energy (Eq.~\ref{eq:ind_fe})
  & 0.82\spc/\spc1.28 & 0.34\spc/\spc0.62 & 0.49\spc/\spc0.86 & 0.57\spc/\spc1.06 & 0.39\spc/\spc0.70 & 0.52\spc/\spc0.90 \\
  
  & \textcircled{\small 2} w/o Goal Supervision (Eq.~\ref{eq:goal_sup})
  & \underline{0.40}\spc/\spc0.61 & 0.15\spc/\spc0.21 & \underline{0.25}\spc/\spc\underline{0.43} & 0.20\spc/\spc \underline{0.35} & 0.22\spc/\spc0.36 & 0.24\spc/\spc0.39 \\
  
  & \textcircled{\small 3} w/o Social Free Energy (Eq.~\ref{eq:social})
  & \underline{0.40}\spc/\spc0.63 & 0.17\spc/\spc 0.24 & 0.27\spc/\spc0.47 & 0.25\spc/\spc0.43 & 0.23\spc/\spc0.39 & 0.26\spc/\spc0.43 \\

  \hline

\multirow{2}{*}{\shortstack{Trajectory \\ Generator}}
  & \textcircled{\small 4} w/o Token-level Condition (Eq.~\ref{eq:token_input})
  & 0.46\spc/\spc0.74 & 0.16\spc/\spc0.23 & 0.30\spc/\spc0.49 & 0.21\spc/\spc\underline{0.35} & 0.18\spc/\spc0.30 & 0.26\spc/\spc0.42 \\
  
  & \textcircled{\small 5} w/o Diffusion Module
  & \underline{0.40}\spc/\spc\underline{0.57} & \underline{0.13}\spc/\spc\underline{0.20} & \underline{0.25}\spc/\spc0.44 & \underline{0.17}\spc/\spc\textbf{0.30} & \underline{0.15}\spc/\spc\underline{0.25} & \underline{0.22}\spc/\spc\underline{0.35} \\

  \hline
  \rowcolor{lightblue}
  & \textcircled{\small 6} \textbf{\model\ (Full)}
 &\textbf{0.37}\spc/\spc\textbf{0.51}  &\textbf{0.12}\spc/\spc\textbf{0.18} &\textbf{0.22}\spc/\spc\textbf{0.39} &\textbf{0.16}\spc/\spc\textbf{0.30}&\textbf{0.14}\spc/\spc\textbf{0.24} &\textbf{0.20}\spc/\spc\textbf{0.32}  \\
  \hline
  \end{tabular}
  \end{table*}

\subsubsection{Efficiency Comparison}
Table~\ref{tab:efficiency} presents the efficiency comparison with state-of-the-art methods. \model\ achieves the best ADE/FDE performance with \textbf{0.20}/\textbf{0.32} at \textbf{4.106}M ($0.926$M for the \textit{Belief Learner} and $3.18$M for the \textit{Trajectory Generator}) parameters and an inference time of \textbf{15.16} ms/agent, suitable for real-time deployment.
While lightweight methods such as EigenTrajectory~\cite{bae2023eigentrajectory} ($0.022$M, $1.425$ ms/agent) and NPSN~\cite{bae2022non} ($0.031$M, $0.414$ ms/agent) offer lower parameter counts and faster inference, they rely on global 
observations and yield inferior prediction accuracy. 
In contrast, \model\ operates under agent-centric partial observability and still achieves superior performance, demonstrating that full global
observations are not a prerequisite for accurate trajectory prediction and that
reasoning from each agent's local observations offers a more practical
formulation for realistic deployment.

\begin{table}[htbp]
\centering
\normalsize
\setlength{\tabcolsep}{2pt}
\renewcommand{\arraystretch}{0.8}
\caption{Efficiency comparison with state-of-the-art methods, averaged across all five scenes. Params are reported in millions (M), inference time in ms/agent, and prediction performance in average minADE$_{20}$/minFDE$_{20}$ (lower is better).}
\setlength{\tabcolsep}{6pt}
\label{tab:efficiency}
\begin{tabular}{lccc}
\toprule
Method & Params & Inference & ADE\spc/\spc FDE \\
\midrule
AgentFormer~\cite{yuan2021agentformer}          & 0.592    & 18.07    & 0.23\spc/\spc0.39 \\
EigenTrajectory~\cite{bae2023eigentrajectory}   & \textbf{0.022}    & 1.425    & 0.22\spc/\spc0.36 \\
GP-Graph~\cite{bae2022learning}                 & 0.036    & 8.423    & 0.24\spc/\spc0.39 \\
NPSN~\cite{bae2022non}                          & \underline{0.031}    &\underline{0.414}    & \underline{0.21}\spc/\spc0.36 \\
Trajectron++~\cite{salzmann2020trajectron++}    & 0.128    & 1.241    & 0.34\spc/\spc0.56\\
SingularTraj~\cite{bae2024singulartrajectory}   & 1.933    &\textbf{0.184}    & \underline{0.21}\spc/\spc\textbf{0.32} \\
MID~\cite{gu2022stochastic}  &9.651 &209.4 &0.25\spc/\spc0.49  \\
SocialVAE~\cite{xu2022socialvae} &2.144 &15.20 &0.22\spc/\spc0.35\\
MoFlow~\cite{fu2025moflow}   &4.701 &1.392 &\underline{0.21}\spc/\spc\underline{0.34}\\
\rowcolor{lightblue}
\textbf{\model\ (ours)}                      &4.106   &15.16 &\textbf{0.20}\spc/\spc\textbf{0.32} \\
\bottomrule
\end{tabular}
\end{table}

\subsubsection{Ablation Studies}
We evaluate the contribution of each component in \model\ by removing it individually while keeping all others intact, as presented in Table~\ref{tab:ablation}.
The \textit{Belief Learner} comprises three components: 
individual free energy (Eq.~\ref{eq:ind_fe}),  goal supervision (Eq.~\ref{eq:goal_sup}), and social free energy (Eq.~\ref{eq:social}),
ablated as variants~\textcircled{\small 1}--\textcircled{\small 3}.
The \textit{Trajectory Generator} is ablated by removing the token-level
condition and the diffusion module, corresponding to
variants~\textcircled{\small 4} and~\textcircled{\small 5}.
In Table~\ref{tab:ablation}, removing any single component consistently degrades performance, validating
the necessity of each design choice. Specifically, variant~\textcircled{\small 1} leads to a substantial performance drop from $\textbf{0.20}/\textbf{0.32}$ to $\textbf{0.52}/\textbf{0.90}$, indicating that individual free energy (Eq.~\ref{eq:ind_fe}) is critical for grounding each agent's belief in its own motion dynamics.
Notably, variants~\textcircled{\small 2} and~\textcircled{\small 3} lead to significant performance degradation, underscoring the indispensable role of goal supervision and social free energy in maintaining stable and accurate predictions.
Variant \textcircled{\small 4} degrades performance from $\textbf{0.20}/\textbf{0.32}$ to
$\textbf{0.26}/\textbf{0.42}$, demonstrating the effectiveness of token-level conditioning in
guiding the diffusion process.
Variant \textcircled{\small 5} yields $\textbf{0.22}/\textbf{0.35}$, indicating that learning belief representations alone is insufficient; cognitively grounded decoding of
agent beliefs into structured tokens is essential for accurate trajectory prediction.

\begin{figure}[htbp]
    \centering
\includegraphics[height=4.7cm,width=8.6cm]{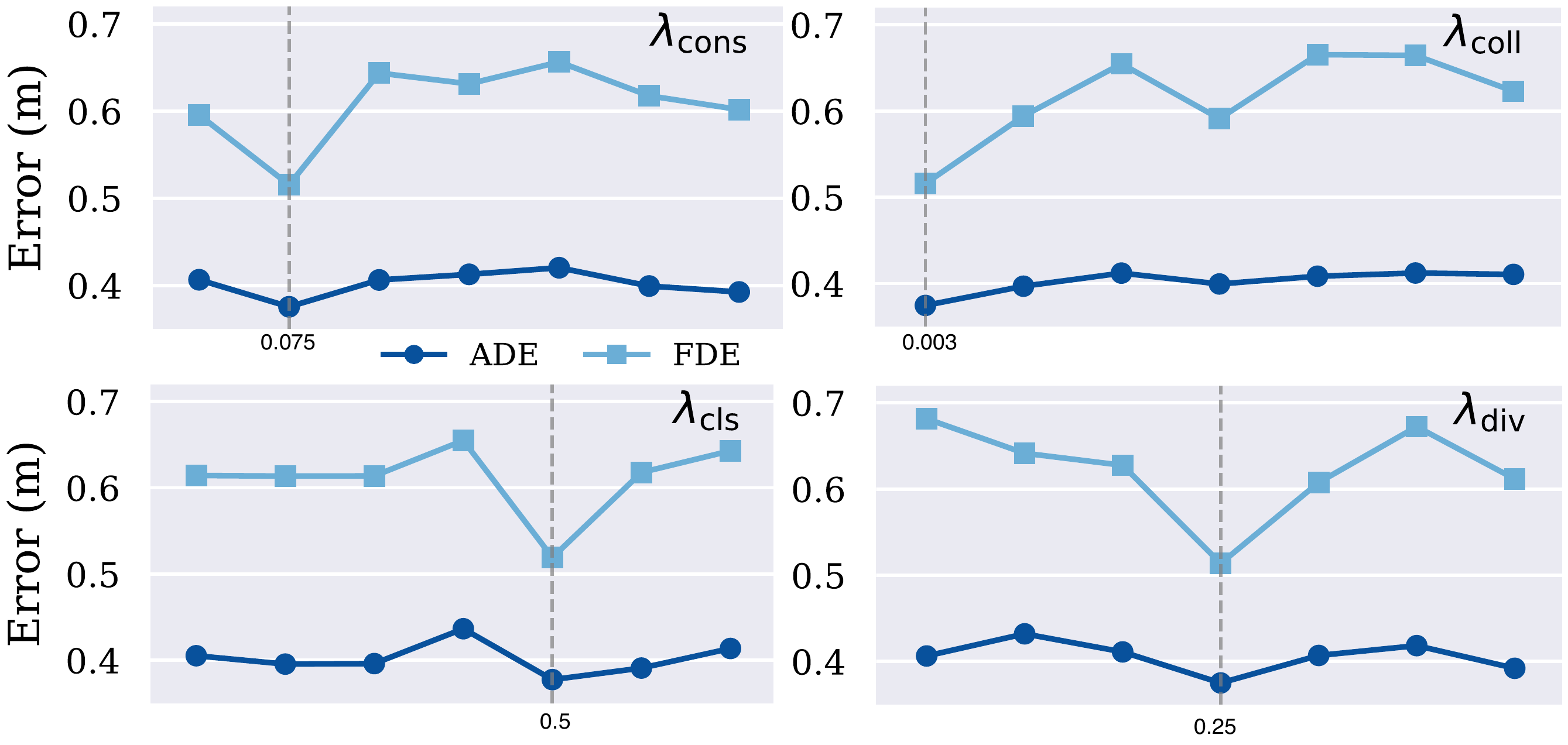}
    \caption{Sensitivity analysis of loss weights on ETH.}
    \label{fig:sensitivity}
\end{figure}

\subsubsection{Parameter Sensitivity Analysis}
Figure~\ref{fig:sensitivity} presents the sensitivity analysis of four loss
weights ($\lambda_{\mathtt{cons}}$, $\lambda_{\mathtt{coll}}$,
$\lambda_{\mathtt{cls}}$, $\lambda_{\mathtt{div}}$) on the ETH scene.
For each weight, we vary its value over a range while keeping all others fixed
at their default settings (dashed lines: $\lambda_{\mathtt{cons}}=0.075$,
$\lambda_{\mathtt{coll}}=0.003$, $\lambda_{\mathtt{cls}}=0.5$,
$\lambda_{\mathtt{div}}=0.25$).
Overall, \model\ is robust to moderate variations in all four weights, with
ADE and FDE remaining stable across a wide range of values around the optimal.
Performance degrades consistently when values deviate substantially from the default, confirming the necessity of each loss term.
For instance, $\lambda_{\mathtt{div}}$ achieves the best minADE$_{20}$/minFDE$_{20}$ of
\textbf{0.37}/\textbf{0.51} at the default value of $0.25$, while doubling it to $0.5$
leads to a noticeable degradation.

\section{Conclusion}

In this work, we present \model, a novel agent-centric trajectory prediction framework grounded in the Free Energy Principle, designed to address the unrealistic global-state assumption in real-world deployment.
\model\ integrates three key components: a dual-branch spatiotemporal encoder that extracts ego-motion and social interaction cues from local observations, a goal-conditioned belief learner that infers latent beliefs with neighborhood consistency constraints, and a residual diffusion trajectory generator with token-level proxy conditioning that produces precise and diverse future predictions.
Extensive experiments on five public benchmarks demonstrate that \model\ consistently outperforms state-of-the-art methods, validating the effectiveness of reasoning from local observations for cognitively consistent trajectory prediction.

\bibliographystyle{ACM-Reference-Format}
\bibliography{references}


\end{document}